\documentclass[twoside,11pt]{article}

\usepackage{jmlr2e}
\usepackage{inconsolata}
\usepackage{pythonhighlight}

\newcommand{\kmlib}{\texttt{kernelmethods}~}
\newcommand{\KM}{\texttt{KernelMatrix}~}
\newcommand{\KS}{\texttt{KernelSet}~}
\newcommand{\KB}{\texttt{KernelBucket}~}

\jmlrheading{1}{2019}{}{12/19}{}{raamana19a}{Pradeep Reddy Raamana}

\ShortHeadings{kernel methods python library}{Raamana}
\firstpageno{1}

\begin{document}
\title{\texttt{kernelmethods} for pattern analysis and machine learning in python}
\author{\name Pradeep Reddy Raamana \email praamana@research.baycrest.org \\
       \addr Rotman Research Institute\\ 
       Baycrest Health Sciences\\
       Toronto, ON, M6A 2E1, Canada
       }
\editor{}

\maketitle

\begin{abstract}%   <- trailing '%' for backward compatibility of .sty file
Kernel methods have proven to be powerful techniques for pattern analysis and machine learning (ML) in a variety of domains. However, many of their original or advanced implementations remain in Matlab. With the incredible rise and adoption of Python in the ML and data science world, there is a clear need for a well-defined library that enables not only the use of popular kernels, but also allows easy definition of customized kernels to fine-tune them for diverse applications. The \kmlib library fills that important void in the python ML ecosystem in a domain-agnostic fashion, allowing the sample data type to be anything from numerical, categorical, graphs or a combination of them. In addition, this library provides a number of well-defined classes to make various kernel-based operations efficient (for large scale datasets), modular (for ease of domain adaptation), and inter-operable (across different ecosystems). %The primary focus of this library is to enable the user to easily define custom kernel functions, as well as enable them to develop advanced techniques like multiple kernel learning to fit their diverse applications. 
The library is available at \url{https://github.com/raamana/kernelmethods}.
\end{abstract}

\section{Introduction}
% \citep{}

% points to be made:
%	kernel methods offer a powerful framework - demonstrated success in many fields
Kernel methods (KM) demonstrated their potential in pattern analysis and machine learning (ML) in a diverse array of domains such as image processing, text analysis, bioinformatics and medicine \citep{shawe2004kernelmethodsbook}. Much of their power and popularity boils down to the kernel trick and the modularity of the kernel learning algorithms. The kernel functions enable the study of non-linear relations in the high-dimensional data by efficiently embedding them in an inner-product feature-space. Kernel-learning algorithms operate exclusively on the the pairwise inner products (so called gram- or kernel-matrix) without needing to know the original representation. This modularity led to efficient and robust algorithms such as the support vector machines (SVM) that performed well in diverse ML applications \citep{burges1998tutorial,vapnik2000statlearning}.

With rapid growth and adoption of open source software, ML libraries have become available in most languages, including but not limited to Java, Python, C++, and Matlab \citep{MLOSS}. Although these languages have varying levels of support for kernel methods, basic functionality of standard kernel functions and SVMs is generally available in many popular ML libraries \citep{svmsdotorg}. However, to fully leverage the power of kernel methods, researchers need to be able to develop, use and fine-tune custom kernel functions and algorithms for their unique application, dataset or domain. But unfortunately many existing libraries are neither easily modifiable nor extensible, and allow limited functionality typically restricted to using predefined kernel functions and algorithms. There have been attempts to address this e.g. \texttt{JKernelMachines} and \texttt{KELP} in Java \citep{picard2013jkernelmachines,filice2017kelp}, and \texttt{kernlab} in R \citep{karatzoglou2004kernlab}. But no such library exists in Python, despite its incredible adoption and accelerated growth in ML, data science and research software development in general \citep{Raamana2018python}. Popular libraries like \texttt{scikit-learn} \citep{pedregosa2011scikitlearn} provide only basic kernel functions, and implementing new kernels or methods is neither easy nor well-defined. This challenge is even more apparent when attempting to implement advanced methods such as multiple kernel learning (MKL) \citep{gonen2011mkl,Raamana2014ThickNetFusion_NBA} and hyper-kernels \citep{Ong2005HyperKernels}. Moreover, common and necessary kernel operations such as normalization, centering and alignment etc are simply missing, in addition to the lack of thorough kernel-oriented validation of such implementations.

\texttt{kernelmethods} aims to fill this important gap in the python ecosystem, to provide a \texttt{scikit-learn}-like library focused on kernel methods, with the following goals: 1) an intuitive API, 2) deep extensibility allowing for easy customization and optimization for diverse needs, and 3) high modularity allowing different data types (from numbers to categorical to graphs to trees) as well as mixed data types (e.g. in medical, epidemiological and biostatistics applications). \texttt{kernelmethods} is a pure python library released under the Apache 2.0 license, implements the best practices in research software engineering, along with continuous integration and a test suite with over 95\% coverage. By filling an important gap, we believe this library would strengthen the python ML ecosystem and advance the state of the art in kernel methods. The source code is available at \url{https://github.com/raamana/kernelmethods} and the corresponding auto-generated documentation is available at \url{https://raamana.github.io/kernelmethods}.

% Given the unique advantages of the python ecosystem (such as such as readability, composable syntax and great community spirit with high rates of open source contributions), and its rapid growth, we believe \texttt{kernelmethods} would enable many users and experts alike leverage the full power of as well as advance the state of the art in kernel methods.

% comparison to JKernelMachine, KELP and kernlab

% runtime profiling? 
% empirical performance!

\section{Design of the \texttt{kernelmethods} library}

\kmlib offers the following features 1) popular kernel functions, 2) \KM class, 3) container classes to manage large collections of kernel matrices, 4) kernel operations and utilities module, and 5) drop-in Estimator classes for ease of use in the python ML ecosystem. It is designed to ease the development of advanced functionality such as customized, composite or hyper kernels.

\subsection{The \KM class}

The \KM is a self-contained class for the Gram matrix induced by a kernel function on a given sample $X$. This class defines the \emph{central} data structure for all kernel methods, as it acts a key bridge between input data space and the learning algorithms \citep{shawe2004kernelmethodsbook}. The library
\begin{itemize}
\setlength\itemsep{0.02em}
\item computes only the elements needed, saving computation and storage
\item supports both callable and attribute access, allowing easy access to partial or random portions of \KM. %Indexing is aimed to be compliant with \numpy as much as possible.
\item allows parallel computation of different parts of \KM to speed up processing of large scale datasets (when $N>50K$)
\item allows user-defined attributes, which is ideal for easy identification among a large collection of KMs from generating, filtering and ranking applications for MKL
\item implements necessary kernel operations such as centering and normalization, which are different from those manipulating regular matrices and
\item exposes several convenience attributes (norms, diagonal and centered versions) and classes for ease of extensibility e.g. \texttt{ConstantKernelMatrix}, \texttt{KernelMatrixPrecomputed}.
\end{itemize}
%This library also provides convenience wrappers:
%\begin{itemize}
%\item \texttt{KernelMatrixPrecomputed} turns a precomputed kernel matrix into a KernelMatrix class with all its attractive properties
%\item \texttt{ConstantKernelMatrix} that provides an efficient definition of a KernelMatrix with a constant everywhere
%\end{itemize}

\subsection{Defining Kernel Functions}
Creating new kernel functions is trivial, which boils down to inheriting from the \texttt{BaseKernelFunction}, implementing the \verb|__call__| method, and giving it a human readable \verb|__str__|. This implementation design, focusing on nothing more than two vectors $x$ and $y$, makes it easy for non-expert users and developers alike to define new and interesting kernels for their unique applications. For example, implementing the entire \texttt{Chi2Kernel} is as simple as:
\begin{python}
class Chi2Kernel(BaseKernelFunction):

    def __init__(self, gamma=1.0):
        super().__init__(name='chi2')
        self.gamma = gamma

    def __call__(self, x, y):
        """Actual implementation of kernel func"""
        # not shown are a lot of checks and validation
        return np.exp(-self.gamma * np.nansum(np.power(x - y, 2) / (x + y)))

    def __str__(self):
        """human readable repr"""
        return "{}(gamma={})".format(self.name, self.gamma)

\end{python}

This frees the users from the burden of having to worry about verifying its implementation is PSD, as all kernel functions are validated to satisfy the Mercer's condition \citep{shawe2004kernelmethodsbook}, which make them interoperable with rest of the kernel machinery. Such abstraction of the implementation and application of the kernel function is exactly how this library becomes domain-agnostic, deferring the handling of the data type down to the data structures holding the features, such as \texttt{pyradigm} \citep{Raamana2017Pyradigm}, and the particular domain-relevant kernel in question. 

\subsection{Utilities}

Besides being able to apply basic kernels on a given sample, this library provides necessary kernel operations, such as normalization, centering, product, alignment evaluation, linear combination and ranking (by various performance metrics) of kernel matrices. While kernel functions are commonly applied on a single sample, this class is designed to allow two samples to be attached with potentially differing number of \texttt{samplets}\footnote{We define the term \texttt{samplet} here to be a single data point in a given sample i.e.\ one row in the feature matrix $X$ of size $n\times p$.}. This is not possible in scikit-learn where kernel implementations are hard-coded for specific definitions.

\subsection{Container classes}

The library also provides \KS and \KB container classes for easy management of a large collection of kernels.  Dealing with a diverse configuration of kernels is necessary for automatic kernel selection and optimization in applications such as Multiple Kernel Learning (MKL), hyper kernel and the like \citep{gonen2011mkl}.

\subsection{Domain agnostic}
Besides the numerical kernels, we designed this library to make it easy to develop categorical, string and graph kernels, owing to its great modularity i.e. feature data-type and iteration of the sample are encapsulated into that particular kernel function and the generic \KM class, and they do not interact with the rest of the library. For example, implementing a categorical kernel function is as simple as:

\begin{python}
class MatchCountKernel(BaseKernelFunction):

    def __init__(self, return_perc=True): # Constructor
        self.return_perc = return_perc
        super().__init__('MatchPerc' if self.return_perc else 'MatchCount')

    def __call__(self, vec_c, vec_d):
        """vec_c, vec_d : array of equal-sized categorical variables"""
        # not shown are a lot of checks and validation
        match_count = np.sum(vec_c==vec_d)
        if self.return_perc:
            return match_count / len(vec_d)
        else:
            return match_count
\end{python}

%with the same attractive properties of intuitive and highly-testable API. In addition to providing native implementation of non-numerical kernels, we aim to provide a deeply and easily extensible framework for arbitrary input data types, such as sequences, trees and graphs etc, via data structures such as pyradigm.

\subsection{Interoperability}

Moreover, drop-in \texttt{Estimator} classes are provided for seamless usage in the scikit-learn ecosystem. For example, SVM with any arbitrary user-defined kernel (on any data type) can be achieved easily with only a few lines of code:

\begin{python}
from kernelmethods import KernelMachine # valid sklearn estimator
from userlib import custom_metric # user-defined kernel function
km = KernelMachine(custom_metric) # learning algorithm is another option
km.fit(X=sample_data, y=targets)  # can be dropped in as estimator anywhere
predicted_y = km.predict(sample_data)
\end{python}

%\subsection{Test suite}

% all the implementations of kernel functions are tested to ensure they are PSD, which is not done at all in other libraries
% 	by checking the kernel matrix induced from them is PSD

% each key and advertised behaviour of the classes comes with an associated set of tests to ensure they work as they are documented.

% all testing is done via continuous integration incorporating the best practices. The test coverage stands at 95%% at the time of submission.

%\section{Extensibility}

\section{Validation}

The \kmlib library is tested thoroughly under continuous integration with over 95\% test coverage. In addition, we run few experiments here to not only demonstrate its utility but also to serve as an additional validation at the level of the library. Following \citep{picard2013jkernelmachines}, we evaluate this library, as well as scikit-learn, on 4 datasets from the UCI repo using the same algorithms and parameters (repeated holdout cross-validation, 80\% training, 20 repetitions). The accuracy estimates for 4 libraries for the default SMO option (gaussian kernel with $\sigma=0.1$) are shown in Table \ref{table:perf}. This demonstrates the \kmlib is performing just as well as the other libraries as expected. The minor differences across libraries are due to differences in implementation of the formulae and parameter interpretation. 

\begin{table}[htp]
\caption{Accuracy estimates of SVM from \kmlib, Weka and JKernelMachines on UCI data sets, which perform similarly, as expected.}
\begin{center}
\begin{tabular}{|r|c|c|c|c|}
\hline
Dataset (N x p) & Weka & JKernelMachines & scikit-learn & \kmlib \\
\hline
ionosphere (351x34)    & 0.861 $\pm$ 0.041 & 0.915 $\pm$ 0.025 & 0.9465 $\pm$ 0.027 & 0.9451 $\pm$ 0.0263\\
heart   (270x13)  & 0.840 $\pm$ 0.038 & 0.831 $\pm$ 0.043 & 0.8167 $\pm$ 0.050 & 0.8037 $\pm$ 0.0348 \\
breast-cancer (638x10) & 0.973 $\pm$ 0.013 & 0.942 $\pm$ 0.018 & 0.9745 $\pm$ 0.011 & 0.9704 $\pm$ 0.0149 \\
german (1000x24) & 0.754 $\pm$ 0.023 & 0.689 $\pm$ 0.029 & 0.7583 $\pm$ 0.026 & 0.7282 $\pm$ 0.0274 \\
\hline
\end{tabular}
\end{center}
\label{table:perf}
\end{table}%

With further tuning (dataset-specific kernel function and parameters), the performance can be improved. As such an exercise would be domain-specific, broader discussion (of dataset properties, performance optimization and trade-offs) is beyond the scope of this software paper.

\subsection{Future work}
We plan to extend the \kmlib library in many important directions: 1) providing base classes for MKL and  HyperKernel (as well as their popular variations) to enable the user focus on higher-level domain-specific optimizations, 2) support for missing data and covariates (when estimating similarity), and 3) supporting broader domains such as graphs, trees and sequences, which have exciting applications in most biomedical research domains.

% Acknowledgements should go at the end, before appendices and references

\acks{PRR would like to acknowledge support for this project from the Canadian Open Neuroscience Program (CONP), and the Ontario Neurodegenerative Disease Research Initiative (ONDRI) and the Canadian Biomarker Integration Network for Depression (CANBIND) programs of the Ontario Brain Institute (OBI).}

%\textbf{Dedication} This library is dedicated to \href{https://en.wikipedia.org/wiki/The_Concert_for_Bangladesh}{The Concert for Bangladesh}, George Harrison and Pandit Ravi Shankar, who moved me immensely with their empathy and kindness, by organizing a benefit concert ever to raise international awareness and funds for Bangladesh's liberation war in 1971.


\newpage

%\appendix
%\section*{Appendix A.}
%\label{app:theorem}
%
%% Note: in this sample, the section number is hard-coded in. Following
%% proper LaTeX conventions, it should properly be coded as a reference:

\vskip 0.2in
\bibliography{references}

\end{document}